\documentclass[runningheads]{llncs}

\usepackage{graphicx}

\usepackage[english]{babel}
\usepackage[utf8]{inputenc}
\usepackage{epstopdf}
\usepackage{url}
\usepackage[bottom]{footmisc}

\usepackage{epstopdf}
\usepackage{makeidx}         
\usepackage{graphicx}        
\usepackage{multicol}        

\usepackage{amsmath,bm,amsfonts,amssymb}

\usepackage{commath}      
\usepackage{subfig}
\usepackage{subfloat}
\usepackage{color,xcolor,ucs}
\usepackage[font=small,labelfont=bf]{caption}

\usepackage{floatrow}
\usepackage{tabularx}
\usepackage{float}
\usepackage{amsthm}

\usepackage{cite}
\usepackage{xr-hyper}
\usepackage{wrapfig}
\usepackage[ruled,vlined,linesnumbered]{algorithm2e}

\usepackage{multirow}
\usepackage{listings}
\usepackage{lipsum}
\usepackage{textcomp} 
 \SetKwInOut{Input}{require}
 \SetKwInOut{Output}{output}

\newtheorem{defn}{Definition}

\begin{document}
\title{Task Allocation for Multi-Robot Task and Motion Planning: a case for Object Picking in Cluttered Workspaces}
\titlerunning{Task Allocation for Multi-Robot Task and Motion Planning}
%
\vspace{-0.6cm}
\author{Hossein Karami\inst{1} \and
Antony Thomas\inst{1} \and
Fulvio Mastrogiovanni\inst{1}}
\authorrunning{H. Karami et al.}
%
\institute{Department of Informatics, Bioengineering,
Robotics, and Systems Engineering, University of Genoa, Via Opera Pia 13,
16145, Genoa, Italy. \email{hossein.karami@edu.unige.it, antony.thomas@dibris.unige.it, fulvio.mastrogiovanni@unige.it}}
\maketitle              
\vspace{-0.8cm}
\begin{abstract}
We present an AND/OR graph-based, integrated multi-robot task and motion planning approach which 
(i) performs task allocation coordinating the activity of a given number of robots, and 
(ii) is capable of handling tasks which involve an \textit{a priori} unknown number of object re-arrangements, such as those involved in retrieving objects from cluttered workspaces. 
Such situations may arise, for example, in search and rescue scenarios, while locating/picking a cluttered object of interest.
The corresponding problem falls under the category of \textit{planning in clutter}. 
One of the challenges while planning in clutter is that the number of object re-arrangements required to pick the target object is not known beforehand, in general.
Moreover, such tasks can be decomposed in a variety of ways, since different cluttering object re-arrangements are possible to reach the target object. 
In our approach, task allocation and decomposition is achieved by maximizing a \textit{combined} utility function. 
The allocated tasks are performed by an integrated task and motion planner, which is robust to the requirement of an unknown number of re-arrangement tasks.
We demonstrate our results with experiments in simulation on two Franka Emika manipulators.
\vspace{-0.2cm}
\end{abstract}



\section{Introduction}
\label{sec:intro}
\vspace{-0.2cm}
Humans trivially perform complex manipulation tasks, such as picking up a tool from a cluttered toolbox, or grabbing a book from a shelf by re-arranging occluding objects. 
For us, as humans, these tasks seem to be routine, and they do not require much thought. 
Yet, for robots, this is not the case. 
Such complex manipulation tasks require symbolic reasoning to decide which objects to re-arrange so as to reach a target object with motion planning to account for the geometric feasibility of the discrete actions. 
This interaction between symbolic reasoning and motion planning is the subject of integrated Task and Motion Planning (TMP).
Single-robot TMP has been an area of active research. 
Yet, current approaches do not naturally account for the capabilities afforded by the presence of multiple robots, such as tasks that can be decomposed in a variety of ways, task allocation involving many robots, or collision avoidance among robots and between each robot and the workspace. 
A naive extension of single-robot TMP approaches to the multi-robot case would have to treat the multi-robot system as a combined set of single-robot system, which becomes computationally intractable as the number of robots increases. 

TMP finds applications in a variety of areas. 
In search and rescue, locating/picking an object of interest may require re-arranging cluttering objects in debris. 
This presents two main challenges:
(i) the amount of debris to be re-arranged before being able to reach and pick the target object is unknown beforehand, 
(ii) the debris should be safely re-arranged so that it poses no further impediment to reaching the target. 
In this paper, we address the above two challenges from a multi-robot perspective, that is, how multiple robots can interact to achieve reaching a target object in clutter while (i) and (ii) hold.
As for challenge (i), in scenarios such as search and rescue, one may not always be interested in finding optimal solutions. Therefore, currently we are not concerned with optimality in terms of the number of object re-arrangements or load-balancing among the involved robots.
To tackle challenge (ii), in this work we assume the availability of a safe region where the objects to be re-arranged are placed so that they pose no further challenge to reach the target object.

\begin{figure}[t]
\centering
\includegraphics[trim=20cm 5cm 30cm 4cm, clip=true,width = 0.4\textwidth,keepaspectratio=true]{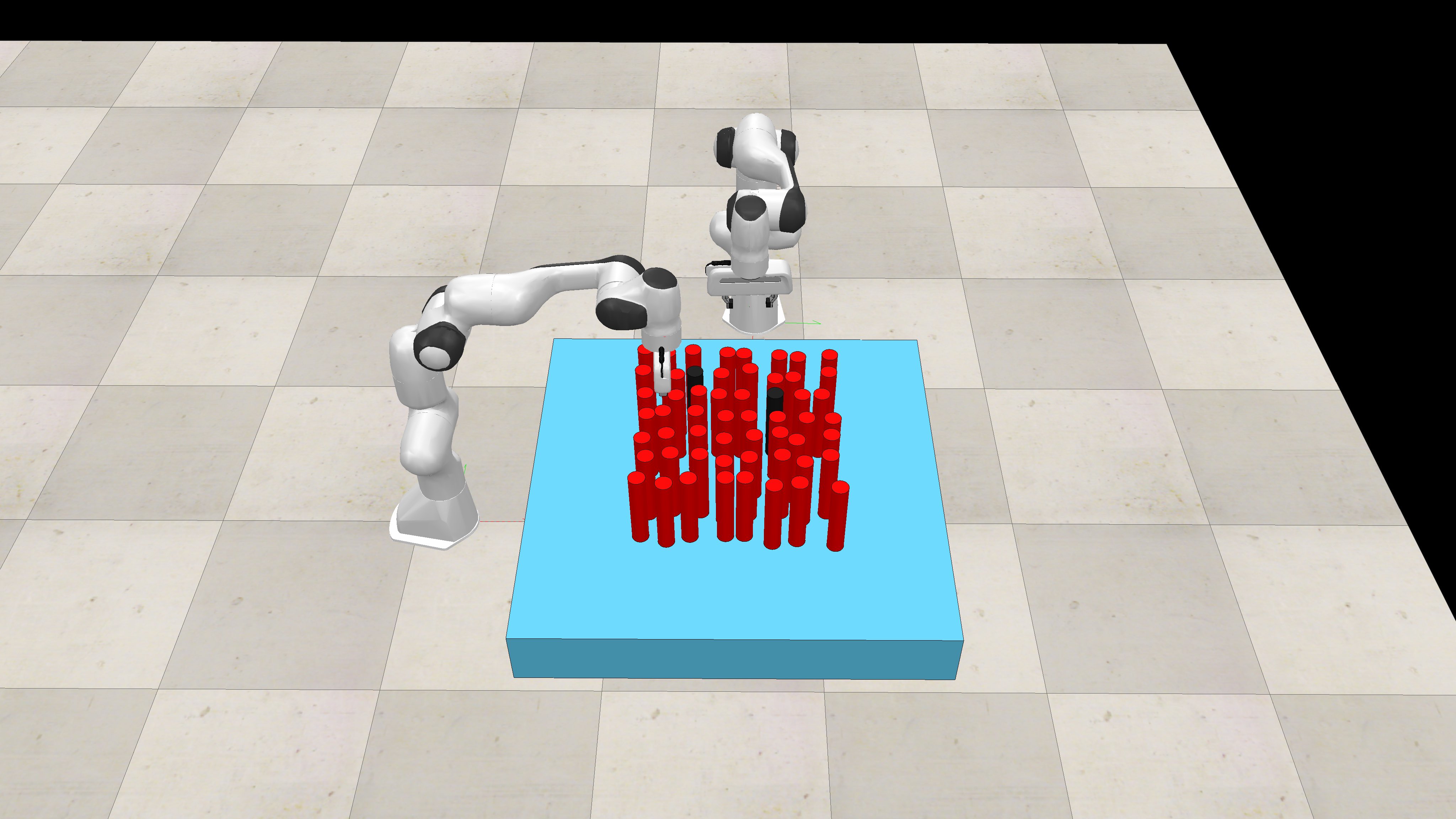}
\caption{Cluttered table-top with two target objects (in black) and other objects in red. 
The multi-robot system consists of two Franka Emika manipulators.
The target objects are allocated to each robot or a single robot depending on the utility function. 
Each target object is then picked by re-arranging clutter via a TMP method.}
\label{fig:scenario}
\end{figure}

We present an approach for multi-robot integrated task and motion planning which first allocates tasks to the available robots, and then plans via AND/OR graphs a sequence of actions for the multiply decomposable tasks that are optimal with respect to an available utility function for the multi-robot system. As a case in point, we focus on re-arranging clutter in manipulation tasks using multiple manipulators. 
Specifically, we consider a cluttered table-top where different target objects are to be picked by the robots by re-arranging occluding objects, as in Fig.~\ref{fig:scenario}. 
This requires task allocation among the available robots, followed by a TMP method to complete the allocated tasks.
It is noteworthy that to achieve each task, that is, picking a target object, the number of cluttering object re-arrangements is not known beforehand. 
Though off-the-shelf {Planning Domain Definition Language} (PDDL)~\cite{mcdermott1998AIPS} based planners are available for task planning, one needs additional expertise to incorporate task-motion interaction that comply with the state-space search of the planner.
Moreover, integrating multi-robot capabilities still remains a challenge. 
As it will be discussed in Section~\ref{sec:approach}, we address these challenges by encoding the task-level abstractions efficiently and compactly within an AND/OR graph that grows iteratively to a tree.   

\section{Background}
\label{sec:background}
\vspace{-0.2cm}
\subsection{Related Work}
\label{sec:related}
\vspace{-0.2cm}
Single-robot TMP is an area of research which has received considerable attention in the literature, and spanning two main areas--- TMP for manipulation~\cite{erdem2011ICRA,kaelbling2013IJRR, srivastava2014ICRA,toussaint2015IJCAI,dantam2016RSS,garrett2018IJRR} and TMP for navigation~\cite{stilman2008IJRR,munoz2016RAS,lo2018AAMAS,jiang2019IROS,thomas2019ISRR}. 
These approaches combine symbolic-level action selection of task planning with the continuous trajectory generation of motion planning~\cite{lagriffoul2018RAL}. 
Yet, such methods do not consider the implication of having multiple robots in task allocation and collision avoidance, and would have to treat the multi-robot system as a combined single-robot system, which becomes intractable as the number of robots increases. 

Current TMP approaches for multi-arm robot systems focus on coordinated planning strategies, and consider simple pick-and-place or assembly tasks. 
As such, these methods do not scale to complex manipulation tasks~\cite{rodriguez2016IROS,umay2019ROSE}.
TMP for multi-arm robot systems in the context of welding is considered in~\cite{basile2012RCIM}, whereas~\cite{umay2019ROSE} discusses an approach for multi-arm TMP manipulation. 
However, the considered manipulation task is a simple pick-and-place operation involving bringing an object from an initial position to a goal position, where two tables and a cylindrical object form the obstacles. 
A centralized inverse kinematics solver is employed in~\cite{mirrazavi2018IJRR}. Motion planning for a multi-arm surgical robot is presented in~\cite{preda2015IROS} using predefined motion primitives. Yet, a fine tuning of such primitives towards real experimental platforms remains a challenge.   

TMP for multi-robot systems has not been addressed thoroughly, and therefore the literature is not sufficiently developed.
Henkel~\textit{et al.}~\cite{henkel2019IROS} consider multi-robot transportation problems using a Task Conflict-Based Search (TCBS) algorithm. 
Such an approach solves a combined task allocation and path planning problem, but assigns a single sub-task at a time and hence may not scale well to an increased number of robots.
Interaction Templates (IT) for robot interactions during transportation tasks are presented in~\cite{motes2019RAL}. 
The interactions enable handing over payloads from one robot to another, but the method does not take into account the availability of robots and assumes that there is always a robot available for such an handover. 
Thus, while considering many tasks at a time this framework does not fare well since a robot may not be immediately available for an handover. 
A distributed multi-robot TMP method for mobile robot navigation is presented in~\cite{thomas2020STAIRS}. 
However, they define task-level actions for a pair of robots and therefore optimal solutions are available for an even number of tasks, and only suboptimal solutions are returned for an odd number of tasks. 
Motes~\textit{et al.}~\cite{motes2020RAL} present TMP-CBS, a multi-robot TMP approach with sub-task dependencies. 
They employ a CBS method~\cite{sharon2015AI} in the context of transportation tasks. 
Constructing a conflict tree for CBS requires the knowledge of different constraints which depend on the sub-task conflicts, for example, two robots being present at a given location at the same time. 
However, in the table-top scenario considered in this paper, the number of sub-tasks is not known beforehand. 
The capabilities of the discussed methods are summarized in Table~\ref{tab:comp}. 
\begin{table}[t]
\scalebox{0.72}{
\begin{tabular}{|c|c|c|c|c|} 
\hline
Considered  & Task & Task & Motion & Unknown number\\
method      & allocation  & decompostion & planning & of sub-tasks\\
\hline
\hline
TCBS~\cite{henkel2019IROS} &  \checkmark & & &  \\ 
\hline
IT~\cite{motes2019RAL}  & \checkmark & \checkmark & & \\
\hline
\cite{thomas2020STAIRS} & \checkmark& & \checkmark &  \\
\hline
TMP-CBS~\cite{motes2020RAL}  & \checkmark & \checkmark & \checkmark & \\
\hline
Our  & \checkmark & \checkmark & \checkmark & \checkmark \\
\hline
\end{tabular}}
\caption{Comparison of different MR TMP methods.}
\label{tab:comp}
\end{table}



\vspace{-0.2cm}
\subsection{Task-Motion Planning and AND/OR Graphs}
\label{sec:tmp}
Task planning or classical planning is the process of finding a discrete sequence of actions from the current state to a desired goal state~\cite{ghallab2016book}.
\begin{defn}
\label{def:one}
A \textit{task} domain $\Omega$ can be represented as a state transition system and is a tuple $\Omega = \langle S, A, \gamma, s_0, S_g \rangle$ where:
\vspace{-0.2cm}
\begin{itemize}
\item $S$ is a finite set of states;
\item $A$ is a finite set of actions;
\item $\gamma : S \times A \rightarrow S$ such that $s' = \gamma(s, a)$;
\item $s_0 \in S$ is the start state;
\item $S_g \subseteq S$ is the set of goal states.
\end{itemize}
\end{defn}
\begin{defn} 
The task \textit{plan} for a task domain $\Omega$ is the sequence of actions $a_0,\ldots,a_m$ such that $s_{i+1} = \gamma(s_i, a_i)$, for $i = 0,\ldots,m$ and $s_{m+1}$ \textit{satisfies} $S_g$.
\end{defn}
Motion planning finds a sequence of collision free configurations from a given start configuration to a desired goal~\cite{latombe1991robot}.
\begin{defn}
A \textit{motion planning domain} is a tuple $M = \langle C, f, q_0, G \rangle$ where:
\vspace{-0.2cm}
\begin{itemize}
\item $C$ is the configuration space;
\item $f =\{0,1\}$, for collision $f=0$, else $f=1$;
\item $q_0 \in C$ is the initial configuration;
\item $G \in C$ is the set of goal configurations.
\end{itemize}
\end{defn}
\begin{defn}
A motion \textit{plan} for $M$ finds a collision free trajectory in $C$ from $q_0$ to $q_n \in G$ such that $f=1$ for $q_0,...,q_n$. 
Alternatively, a motion \textit{plan} for $M$ is a function of the form $\tau : [0, 1] \rightarrow C_{free}$ such that $\tau(0) = q_0$ and $\tau(1) \in G$, where $C_{free} \subset C$ is the configurations where the robot does
not collide with other objects or itself.
\end{defn}
TMP combines discrete task planning and continuous motion planning to facilitate efficient interaction between the two domains. 
Below we define the TMP problem formally.
\begin{defn}
A \textit{task-motion planning} with task domain $\Omega$ and motion planning domain $M$ is a tuple $\Psi =\langle C, \Omega, \phi, \xi, q_0 \rangle$ where:
\vspace{-0.2cm}
\begin{itemize}
\item $\phi : S  \rightarrow 2^ C$, maps states to the configuration space; 
\item $\xi : A  \rightarrow 2^ C$, maps actions to motion plans.
\end{itemize}
\end{defn}
\begin{defn}The \textit{TMP problem} for the TMP domain $\Psi$ is to find a sequence of discrete actions $a_0,...,a_n$ such that $s_{i+1} = \gamma(s_i, a_i)$, $s_{n+1} \in S_g$ and a corresponding sequence of motion plans $\tau_0,...,\tau_n$ such that for $i = 0,...,n$, it holds that (i) $\tau_i(0) \in \phi(s_i) \ \textrm{and} \ \tau_i(1)  \in \phi(s_{i+1})$, (ii) $\tau_{i+1}(0) = \tau_i(1)$, and (iii) $\tau_i \in \xi(a_i)$. 
\end{defn}
We now provide a brief overview of AND/OR graphs~\cite{darvish2020hierarchical,karami2020task}. 
An AND/OR graph is a graph which represents a problem-solving process~\cite{chang1971AI}.
\begin{defn}
An AND/OR graph $G$ is a directed graph represented by the tuple $G = \langle N,H \rangle$ where:
\vspace{-0.2cm}
\begin{itemize}
\item $N$ is a set of nodes;
\item $H$ is a set of hyper-arcs.
\end{itemize}
\end{defn}
For a given AND/OR graph $G$, we have that $H = \{h_1,\ldots,h_m\}$, where $h_i$ is a many-to-one mapping from a set of child nodes to a parent node. 
In that sense, a hyper-arc induces a logical \textit{and} relationship between the child nodes/states, that is, all the child states should be satisfied to achieve the parent state. 
Similarly, a single parent node can be the co-domain for different hyper-arcs $h_i$. 
These hyper-arcs are in logical \textit{or} with the parent node.
Nodes without any successors or children are called the \textit{terminal} nodes. 
The terminal nodes are either a success node or a failure node.
We now mathematically define an AND/OR graph network, a detailed treatment of which can be found in~\cite{karami2021arxiv}.
\begin{defn}
For an AND/OR graph $G=\langle N,H \rangle$, an augmented AND/OR graph $G^a$ is a directed graph represented by the tuple $G^a = \langle N^a,H^a \rangle$ where:
\vspace{-0.2cm}
\begin{itemize}
\item $N^a = \{N,n^v\}$ with $n_v$ being the virtual node;
\item $H^a = \{H,H^v\}$ with $H^v=\{h^v_i\}_{1 \leq i \leq |H^v|}$.
\end{itemize}
\end{defn}
\begin{defn}
An AND/OR graph network $\Gamma$ is a directed graph $\Gamma = \langle \mathcal{G},T \rangle $ where:
\vspace{-0.4cm}
\begin{itemize}
\item $\mathcal{G} = \{G^a_1,\ldots,G^a_{n'}\}$ is a set of augmented AND/OR graphs $G^a_i$;
\item $T= \{t_1,\ldots,t_{n'-1}\}$ is a set of transitions such that $G^a_{i+1} = t_i(G^a_i), \ 1 \leq i \leq n'-1$;
\end{itemize}
\end{defn}
\noindent where $n'$ is the total number of graphs in the network. 
Alternatively, $n'$ is also the depth of the network.
\vspace{-0.2cm}
\subsection{Problem Definition}
\label{sec:prob_defn}
We present a framework which
(i) allocates $T$ manipulation tasks to a set of $R$ robots, and
(ii) plans for a sequence of motions-actions for each robot to achieve the respective tasks. 
To this end, we consider a multi-robot setting with a cluttered table-top wherein different target objects need to be picked by re-arranging the clutter. 
Thus, each manipulation task corresponds to picking up a target object. 

In order to carry out each task, objects need to be re-arranged -- depending on the clutter \textit{degree} -- and the number of objects to be re-arranged is not known beforehand. 
By representing the task-level abstractions within an AND/OR graph network (see Section~\ref{sec:approach}), we can overcome this issue.

\section{Multi-robot Task and Motion Planning}
\label{sec:approach}
\begin{figure}[t]
\centering
\subfloat[]{\includegraphics[width=0.234\textwidth]{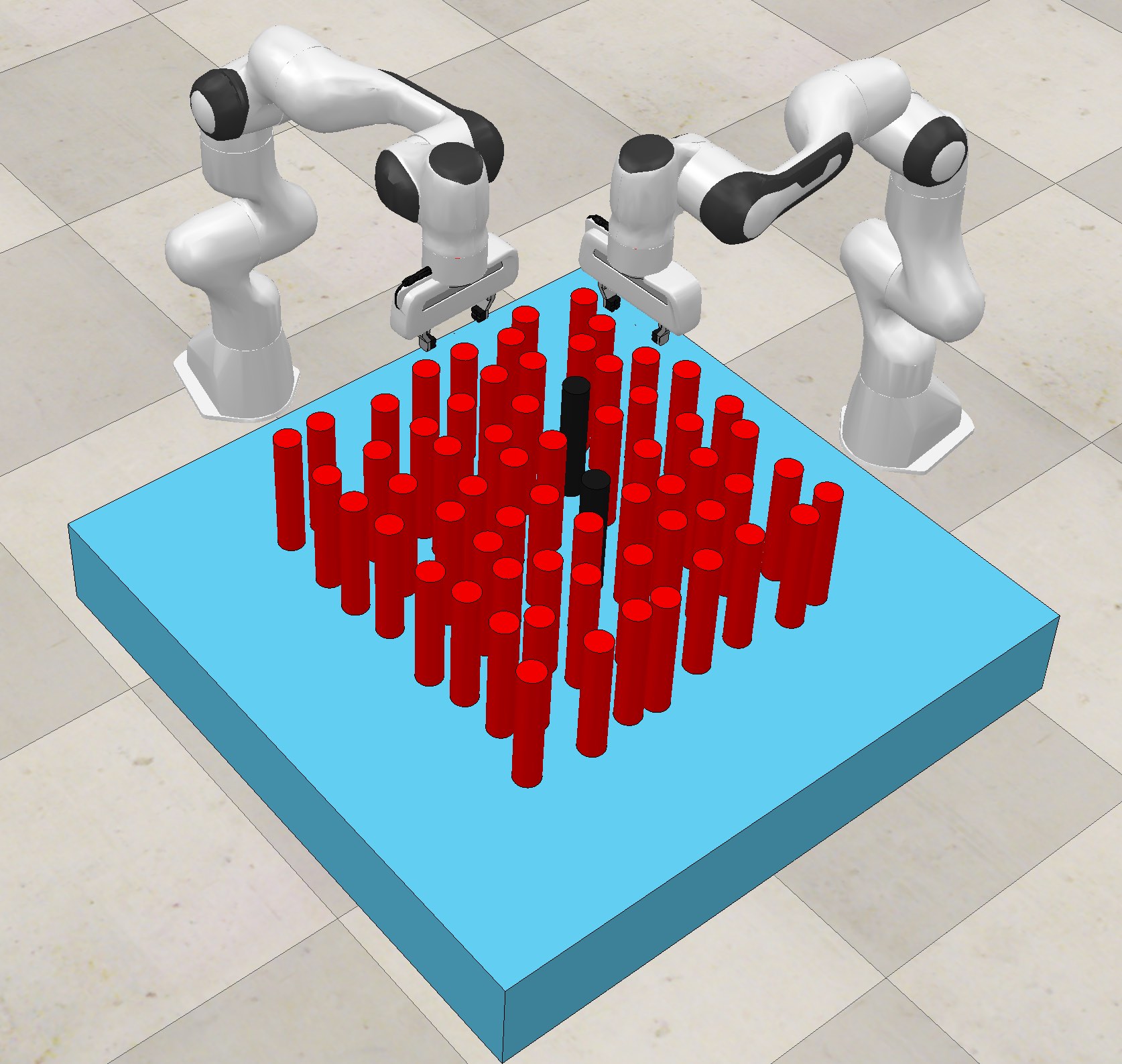}\label{fig:t0}}\hfill
\subfloat[]{\includegraphics[width=0.24\textwidth]{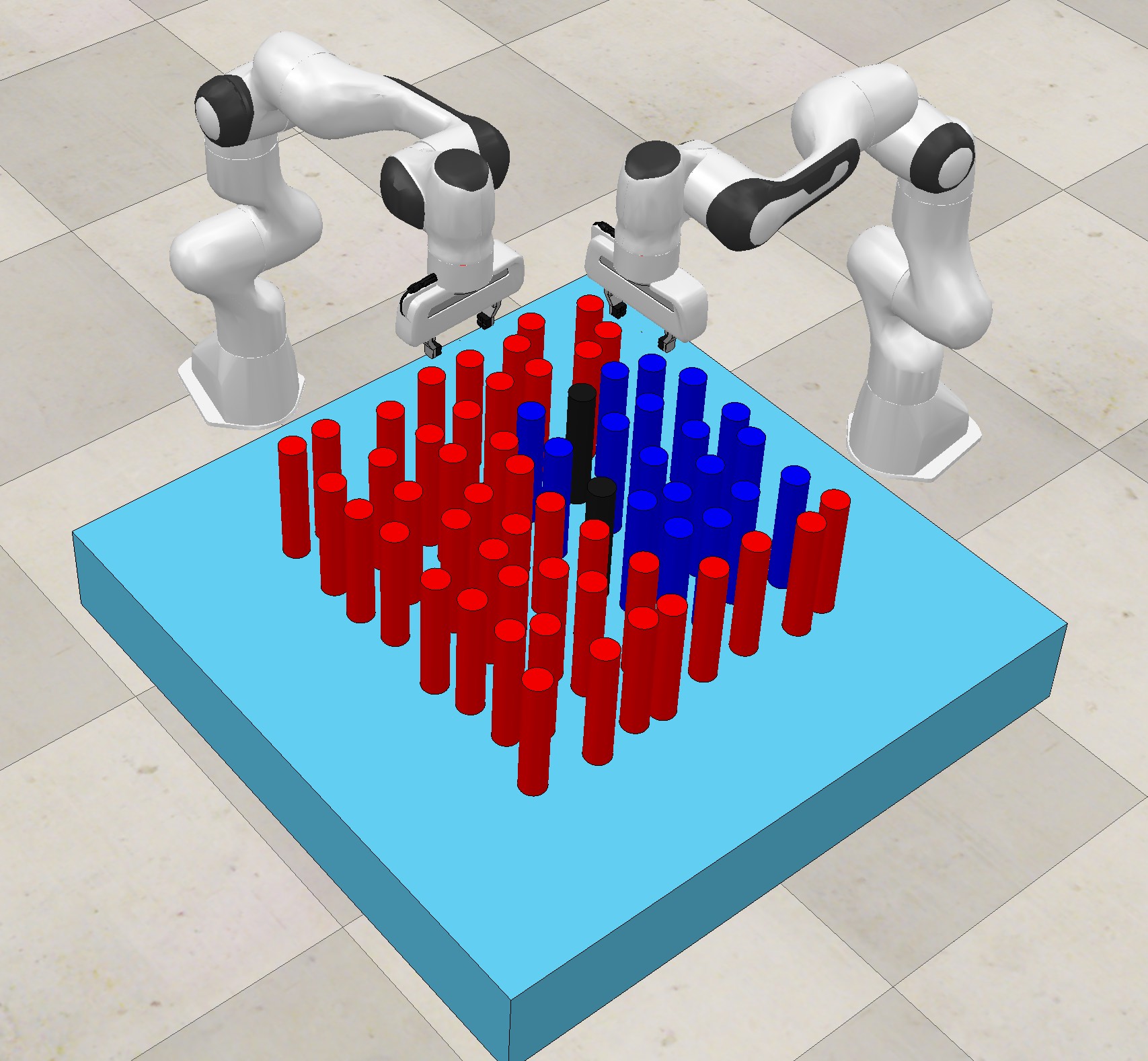}\label{fig:t1}}\hfill
\subfloat[]{\includegraphics[width=0.235\textwidth]{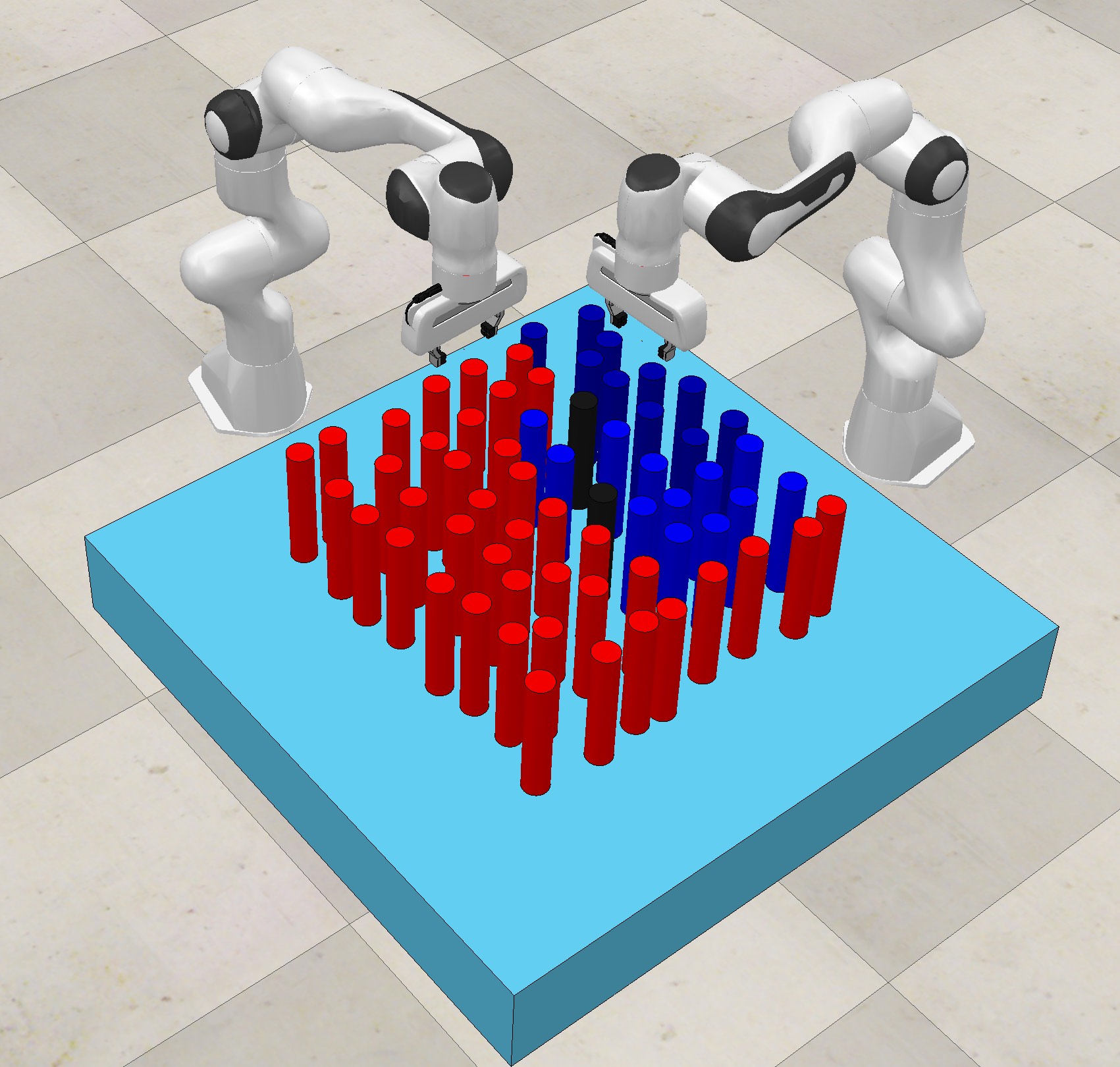}\label{fig:t2}}\hfill
\subfloat[]{\includegraphics[width=0.235\textwidth]{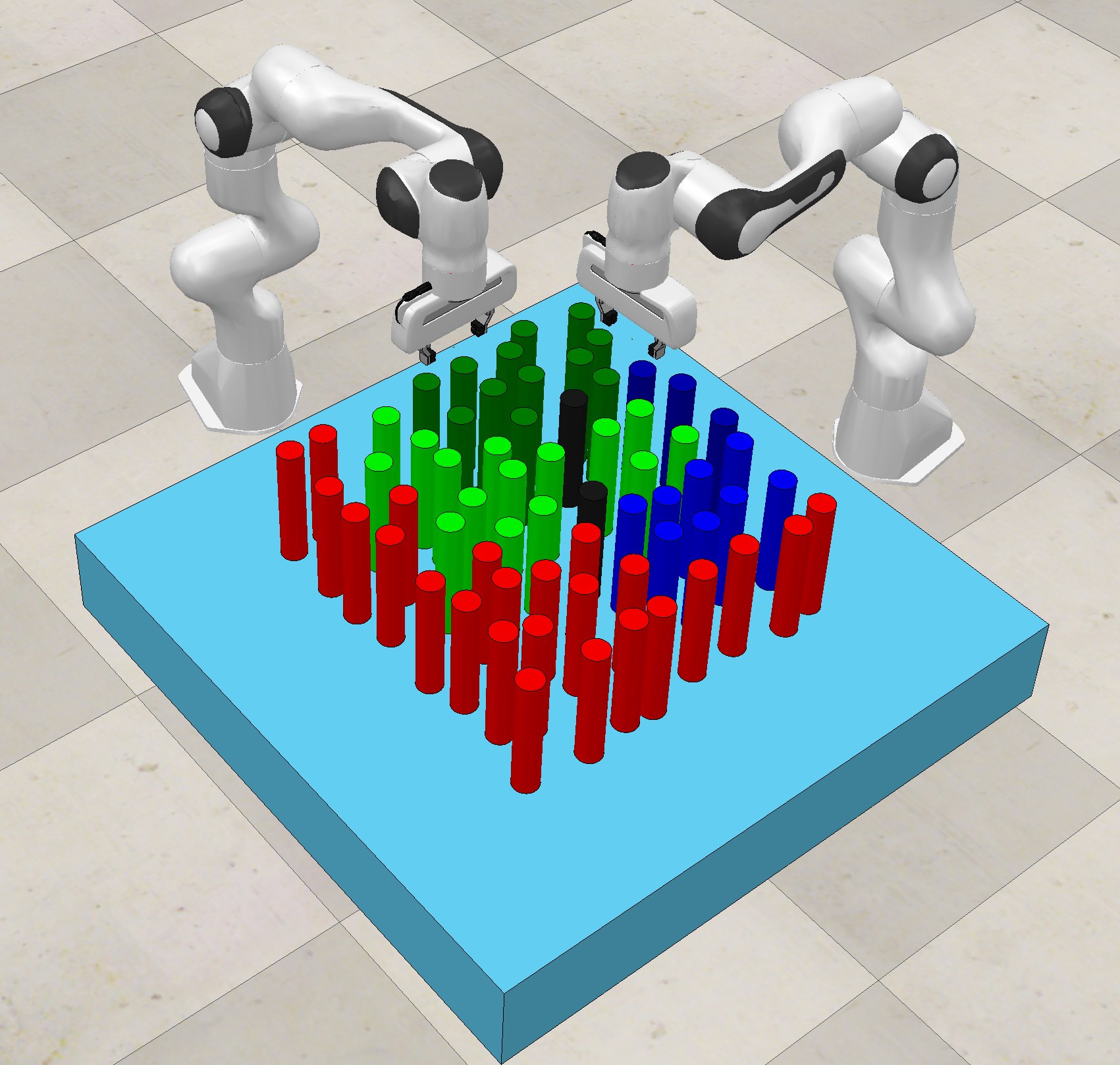}\label{fig:t3}}
\caption{Illustration of the obstacle selection method:
(a) a cluttered table-top scenario with two robots r1 (right), r2 (left) and the target objects in black;
(b) a valid grasping angle range is computed by discretizing a fixed grasping angle range of $-\frac{\pi}{2}$ to $\frac{\pi}{2}$; the objects that fall within the grasping angle range of r1 (one target) are shown in blue;
(c) blue objects within the grasping angle range of r1, considering both the targets;
(d) Objects within the grasping range of both r1 (blue) and r2 (green), considering both the targets for each robot.}
\label{fig:task}
\end{figure}
In this Section, we detail our multi-robot TMP method. 
We begin by describing a heuristic approach providing a rough estimate of the objects to be re-arranged to pick a target object. 
The approach allows us to define a combined utility function for the multi-robot system to perform task allocation. 
Allocated tasks are then carried out using a TMP method based on an AND/OR graph network and a sampling based motion planner.
\vspace{-0.2cm}
\subsection{Obstacles Selection}
An overview of the method can be seen in Figure~\ref{fig:task}.
First, it finds different feasible plans to the target, each one corresponding to different grasping angles, by ignoring all the obstacles in the workspace. 
This is done by discretising the set of graspable angles, that is $\left[-\frac{\pi}{2}, \frac{\pi}{2}\right]$ (the axis is located at the center of the target).
It is noteworthy that we consider only side grasps to make the scenario more challenging. 
Among the available plans, the maximum and minimum grasping angles, namely $\alpha$ and $\beta$, are then obtained. 
The method then constructs two lines starting at the center of the target object and towards the robot with the corresponding angles $\alpha$ and $\beta$. 
The lines are terminated when there are no more obstacles along their paths. 
The end-points are then joined to form a triangle. 
This triangle is then enlarged on all the three sides by the radius of the bounding volume sphere of the end-effector. 
The objects within the constructed triangle are the objects to be re-arranged to facilitate the target grasp. 

We note that what we describe here is an approximate method to identify the objects to be re-arranged.
The actual set of objects depend on other such factors as the degrees of freedom of the robot, the size of the links and the end-effector, or the \textit{degree} of clutter. 
As it will be discussed later, for all practical purposes we are interested only in an approximate measure so as to perform multi-robot task allocation. 
\vspace{-0.4cm}
\subsection{Task Allocation}
Let $R$ be the number of available robots, and $T$ the number of tasks to be allocated, that is, we have $T$ target objects to be picked, such that $T \geq R$. 
Task allocation is performed offline and we assume that each task, that is, picking up a target object from clutter, is performed by a single robot, and that each robot is able to execute only one task at a time. 
We also recall here our assumption from the Introduction that the objects to be re-arranged to reach and pick a target are placed in a \textit{safe} region.
Our task allocation strategy falls under the Single-Task, Single-Robot, Time-extended Assignment (ST-SR-TA) taxonomy of Gerkey and Matari{\'c}~\cite{gerkey2004IJRR}, since the multi-robot system contains more tasks than robots. 
In order to allocate tasks, we define $U_{r_it_j}$ as the utility function for a robot $r_i \in R$ executing a task $t_j \in T$.
In this work, utility is inversely proportional to the number of object re-arrangements required to grasp the target object. 
To determine such a measure, we first (randomly) select a target object $t_1$, and then for each robot $r_i$ we run the obstacles selection algorithm described above. 
For each $r_i$, such a run returns the set of objects to be re-arranged to reach $t_1$. 
Let us denote this set by $O_{r_it_1}$ (and by $O_{r_it_j}$ for the $j$th task). 
The robot $r_k$ whose set $O_{r_kt_1}$ is of minimum cardinality (i.e., maximum utility) is then allocated task $t_1$.
For the next target $t_2$ this step is repeated. 
We note here that $O_{r_kt_2}$ is computed offline and returns the number of objects to be re-arranged by robot $r_k$ to execute task $t_2$. 
However, it may be the case that some objects appear in both $O_{r_kt_1}$ and $O_{r_kt_2}$, that is, $O_{r_kt_1} \cap O_{r_kt_2} \neq \{\emptyset\}$. 
Since each robot executes one task at a time, $r_k$ can execute $t_2$ only after having performed $t_1$. 
Thus, during the execution phase, $r_k$ may have already removed the common objects while executing $t_1$, and therefore these objects may be ignored to avoid \textit{intra-robot} double counting while computing the utility $U_{r_kt_2}$ offline. 

Once each robot is allocated a task, the set of remaining tasks $T'$ is completed only after the execution of the assigned tasks. 
For the remaining $T'$ tasks, the \textit{inter-robot} or robot-robot double counting must be considered. 
Reasoning in a similar manner for intra-robot double counting, the set $O_{r_it_j}$ restricted to $T \setminus T' < j \leq T$ for the remaining $T'$ tasks may have common objects with respect to $O_{r_it_j}$ restricted to $1 < j \leq T'$ of the assigned tasks. 
Thus, the total number of objects to be re-arranged for robot $r_i$ to execute task $t_j$ is
\begin{equation}
O_{r_it_j}^c = |O_{r_it_j}| - \sum_{k} |O_{r_it_jt_{k}}| - \sum_{k} \sum_{l}|O_{r_ir_kt_jt_l}|
\end{equation}
\noindent where $|\cdot|$ denotes the cardinality of a set, 
\begin{equation}
    O_{r_it_jt_{k}} =
    \begin{cases}
    O_{r_it_j}\cap O_{r_it_k} \ &\text{if $r_i$ allotted $t_k$ previously}, \\
    0 \ &\text{otherwise}.
    \end{cases}
\end{equation}
\noindent and
\begin{equation}
    O_{r_ir_kt_jt_l} =
    \begin{cases}
    O_{r_it_j}\cap O_{r_kt_l} \ &\text{if $r_k$ allotted $t_l$ previously}, \\
    0 \ &\text{otherwise},
    \end{cases}
\end{equation}
\noindent with the terms $|O_{r_it_jt_{k}}|$ and $|O_{r_ir_kt_jt_l}|$ modeling the intra-robot and inter-robot double counting, respectively. 
We therefore have the following utility function
\begin{equation}
    U_{r_it_j} = \frac{1}{1+O_{r_it_j}^c}. 
    \label{eq:utility}
\end{equation}
The maximum utility of $U_{r_it_j} = 1$ is therefore achieved when no object re-arrangement is required to execute task $t_j$, that is, $O_{r_it_j}^c = 0$. 
Using the taxonomy in~\cite{korsah2013IJRR}, we thus have In-schedule Dependencies (ID) -- the effective utility of an agent for a task depends on what other tasks that agent is performing as well as Cross-schedule Dependencies (XD) -- the effective utility of an agent for a task depends not only on its own task but also on the tasks of other agents. 

We now define the combined utility of the multi-robot system, which consists of maximising
\begin{equation}
    \sum_{i \in R} \sum_{j \in T} U_{r_it_j}x_{r_it_j}\
    \label{eq:comb_utility}
\end{equation}
\begin{equation}
\begin{split}
& \textrm{such that} \, \sum_{i \in R} x_{r_it_j} = 1\\
& \textrm{where} \, x_{r_it_j} \in \{0,1\}. \\
\end{split}
\label{eq:comb_utility1}
\end{equation}
From~\eqref{eq:utility} and~\eqref{eq:comb_utility} we see that the robot with the minimum number of object re-arrangements for a given task is thus assigned the maximum utility. 
In case of a tie, we select the robot which has not been allocated any task. 
If all the robots with the same utility have been allocated tasks already, or if none has been allotted, then a robot is selected randomly. 
\vspace{-0.2cm}
\subsection{Task Decomposition}
Each manipulation task is decomposed into a set of sub-tasks which correspond to pick-and-place tasks, that is, re-arrangement of the objects that hinder the target grasp. 
As seen above, the number of sub-tasks for a given task are not known beforehand. 
Moreover, \textit{multiple decomposability}~\cite{zlot2006phd} is possible since the clutter can be re-arranged in different ways. 
We seek a decomposition minimizing the number of sub-tasks for the multi-robot system. 
This can be achieved during task allocation since utility is computed based on the obstacles selection method. 
In this work, we consider \textit{complex task decomposition}~\cite{zlot2006phd} -- a multiply decomposable task for which there exists at least one decomposition that is a set of multi-robot allocatable sub-tasks. 
Though in this work we ignore the multi-robot allocatability property, this can be incorporated trivially. 
For example, let us consider the case where two robots have the same utility to perform a task $t_j$. 
In this case, the sub-tasks can be equally divided to achieve multi-robot allocatability or one robot may be selected randomly (or depending on previous allocation) to perform the entire task.
\vspace{-0.4cm}
\subsection{Task and Motion Planning}
\label{sec:task}
PDDL~\cite{mcdermott1998AIPS} is the \textit{de facto} standard for task planning, and most TMP approaches resort to it. 
Integrated TMP requires a mapping between the task space and the motion space.
Semantic attachments are used in~\cite{dornhege2009SSRR, dornhege2009ICAPS, dantam2016RSS, thomas2019ISRR} to associate algorithms to functions and predicate symbols via external procedures.
However, it is assumed by these approaches that the workspace be known in advance. 
Moreover, they reduce motion space to a finite space since the robot configuration, the grasp poses, and other geometrical constraints need to be specified in advance. 
This issue is addressed in~\cite{garrett2020ICAPS} using \textit{streams}, which enable procedures within PDDL to sample values of continuous variables, therefore encoding an infinite set of actions. 
Though off-the-shelf PDDL planners are available, one needs additional expertise to incorporate semantic attachments or streams complying with the planner semantics. 
Furthermore, for the cluttered table-top scenario, the number of objects that need to be re-arranged is not known beforehand. 
As a result, one would need observation based re-planning within the PDDL framework, which require additional mapping between the PDDL action space and the observation space of the robot~\cite{bertolucci2021TPLP}.

A cluttered table-top scenario in the context of single-robot TMP is considered in~\cite{thomas2020IRIM, karami2021arxiv}.  
To address the above mentioned challenges associated with PDDL modeling, in~\cite{karami2021arxiv} the task-level abstractions of the TMP problem is efficiently and compactly encoded within an augmented AND/OR graph (see Section~\ref{sec:background}) that grows iteratively until solutions are found, where each iteration is meant at managing a sub-task as defined above.

In general, AND/OR graphs are constructed offline. 
However, the number of objects to be re-arranged depends on the clutter degree and is not known beforehand.
\cite{karami2021arxiv} introduces AND/OR graph networks, consisting of augmented AND/OR graphs which iteratively deepen at run-time till a solution to the modeled problem is found.
The key idea is that the set of task-level abstractions, i.e., the states (nodes) and actions (hyper-arcs) of the AND/OR graph defined for the robot remain the same irrespective of the number of re-arrangements, i.e., sub-tasks. 
Therefore, one can define an initial AND/OR graph encoding task-level actions, which is augmented with the initial workspace configuration. 
The augmented graph is then iteratively expanded with the updated workspace configuration as long as sub-tasks are carried out resulting in a network of AND/OR graphs.
In this work, we use this AND/OR graph network based task planner and customize it for multi-robot task planning.
For motion planning, we simply use MoveIt~\cite{sucan2013moveit}, which supports RRT~\cite{kuffner2000ICRA} from OMPL~\cite{sucan2012RAM}.
The motion planner is first employed to execute the obstacles selection algorithm.
Once the tasks are allocated, the motion planner is called to (i) achieve the re-arrangement of each sub-task identified by the task planner (i.e., the AND/OR graph network) and (ii) to grasp the target object. 
\vspace{-0.2cm}
\subsection{The Multi-robot Task-Motion Planning Loop}
 \begin{wrapfigure}[26]{l}{0.5\linewidth}
	\centering
	\vspace{-0.8cm}
		\includegraphics[scale=0.25]{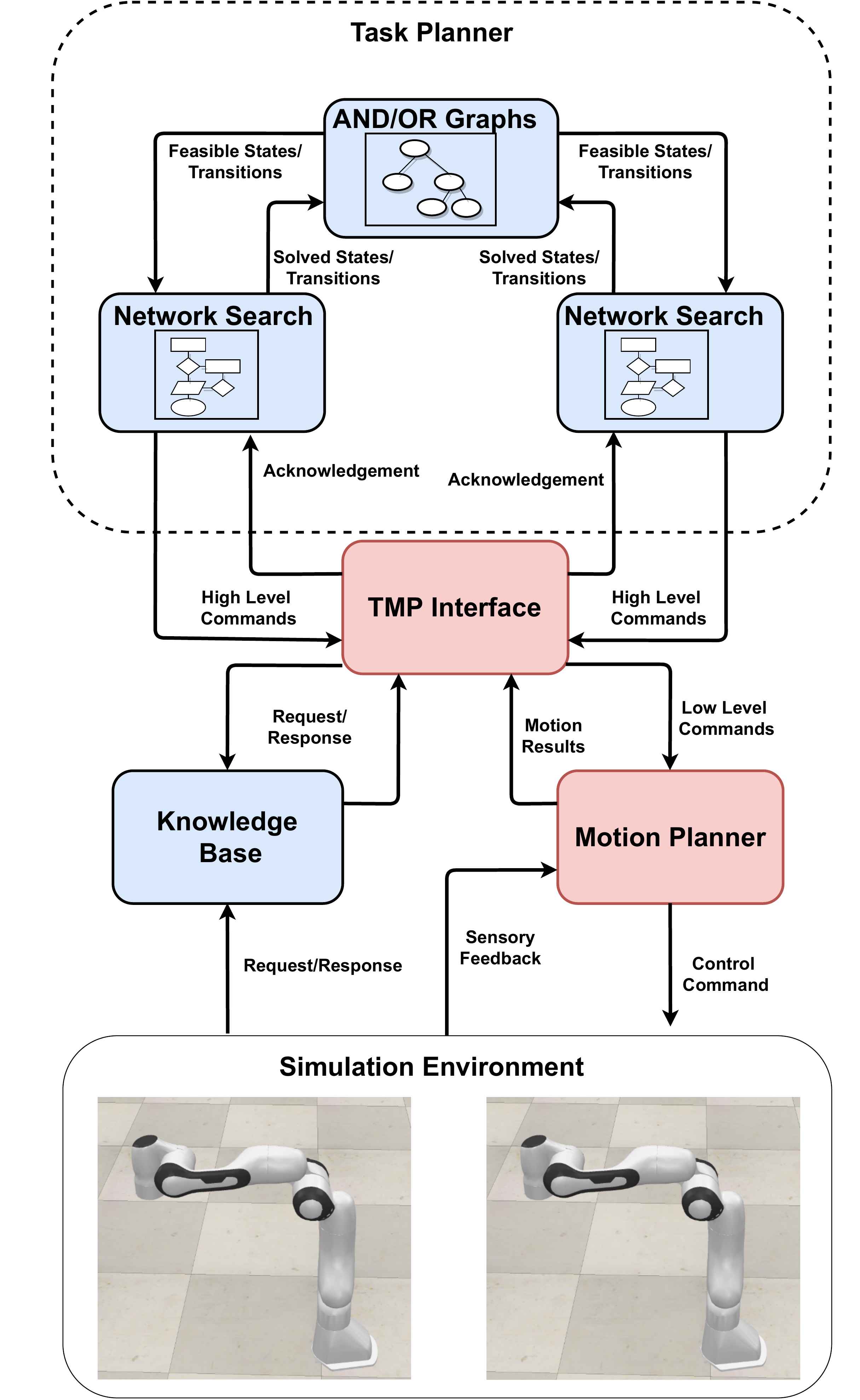}
		\caption{System's architecture.}
	\label{fig:arch}
\end{wrapfigure}
Fig.~\ref{fig:arch} depicts the overall system's architecture of our multi-robot TMP method. 
Once the tasks have been allocated, \textit{Task Planner} selects the abstract actions whose geometric execution feasibility is checked by \textit{Motion Planner}. 
The \textit{Task Planner} layer consists of the \textit{AND/OR Graphs} module -- the initial augmented AND/OR graphs for the robots, and the \textit{Network Search} module -- the search procedure iterating the initial augmented graphs.
As discussed in Section~\ref{sec:task}, the initial augmented AND/OR graph consists of the task-level actions for each robot augmented with the current workspace configuration.
\textit{AND/OR Graphs} provides a set of achievable transitions between the states to \textit{Network Search}, and receives the set of allowed states and transitions as the graph is expanded. 
\textit{Task Planner} then associates each state or state transition with an ordered set of actions in accordance with the workspace and robot configurations. The \textit{Knowledge Base} module stores the information regarding the current workspace configuration, that is, the objects and their locations in the workspace as well as the robot configuration. 
This module augments the graphs with the current workspace configuration to facilitate \textit{Network Search}.
Since we consider here two robots and their respective configurations need to be updated, we have two \textit{Network Search} modules corresponding to each robot. 

\textit{TMP interface} acts as a bridge between the task planning and the motion planning layers. 
It receives action commands from \textit{Task Planner}, converts them to their geometric values (for example, a grasping command requires various geometric values such as the target pose or the robot base pose), and passes them on to \textit{Motion Planner} to check motion feasibility. 
To this end, the module retrieves information regarding both the workspace and robots from \textit{Knowledge Base}. 
If an action is found to be feasible, it is then sent for execution. 
Upon execution, \textit{Task Planner} receives an acknowledgment regarding action completion and the \textit{Knowledge Base} is updated accordingly.

\vspace{-0.3cm}
\section{Experimental Results}
\label{sec:results}
\begin{wraptable}[8]{l}{0.53\linewidth}
\small\sf\centering
\vspace{-0.9cm}
 \scalebox{0.8}{
\begin{tabular}{|l|c|c|}
\hline
Activity                    & Average [s]   & Std. Dev. [s] \\ 
\hline
\hline
AND/OR Graphs               & 0.03213       & 0.02174 \\ 
Network Search              & 0.8992        & 0.1862 \\
Motion planner (attempts)   & 18.320        & 6.450 \\
Motion execution (attempts) & 6.3126        & 0.8176 \\
Motion planner (time)       & 0.8010        & 0.2170 \\
Motion execution (time)     & 4.7490        & 2.0240 \\
\hline
\end{tabular}}
    \caption{Average computation times for different modules.}
      \label{tab:compana}
\end{wraptable} 

We validate and demonstrate the performance of our multi-robot TMP approach by performing experiments in the state-of-the-art robotics simulator CoppeliaSim~\cite{coppeliaSim}, employing two Franka Emika manipulators. 
As seen in Fig.~\ref{fig:scenario}, we consider a cluttered table-top scenario where manipulation tasks correspond to picking up different target objects. 
For each task, the sub-tasks consist of removing objects hindering each target grasp. 
In this work, such objects are picked and placed outside the working area, in a \textit{safe} space.
The number of objects to be removed is unknown (and therefore likewise the number of sub-tasks) at planning time. 
Experiments are conducted on a workstation equipped with an Intel(R) core i7-8700@3.2 GHz $\times$ 12 CPU's and 16 GB of RAM. 
The architecture is developed using C++ and Python under ROS Kinetic. A video demonstrating the results can be found at
\verb|https://youtu.be/nkB_LHorg1g|.
\begin{figure}[t]
\centering
\subfloat[]{\includegraphics[width=0.35\textwidth]{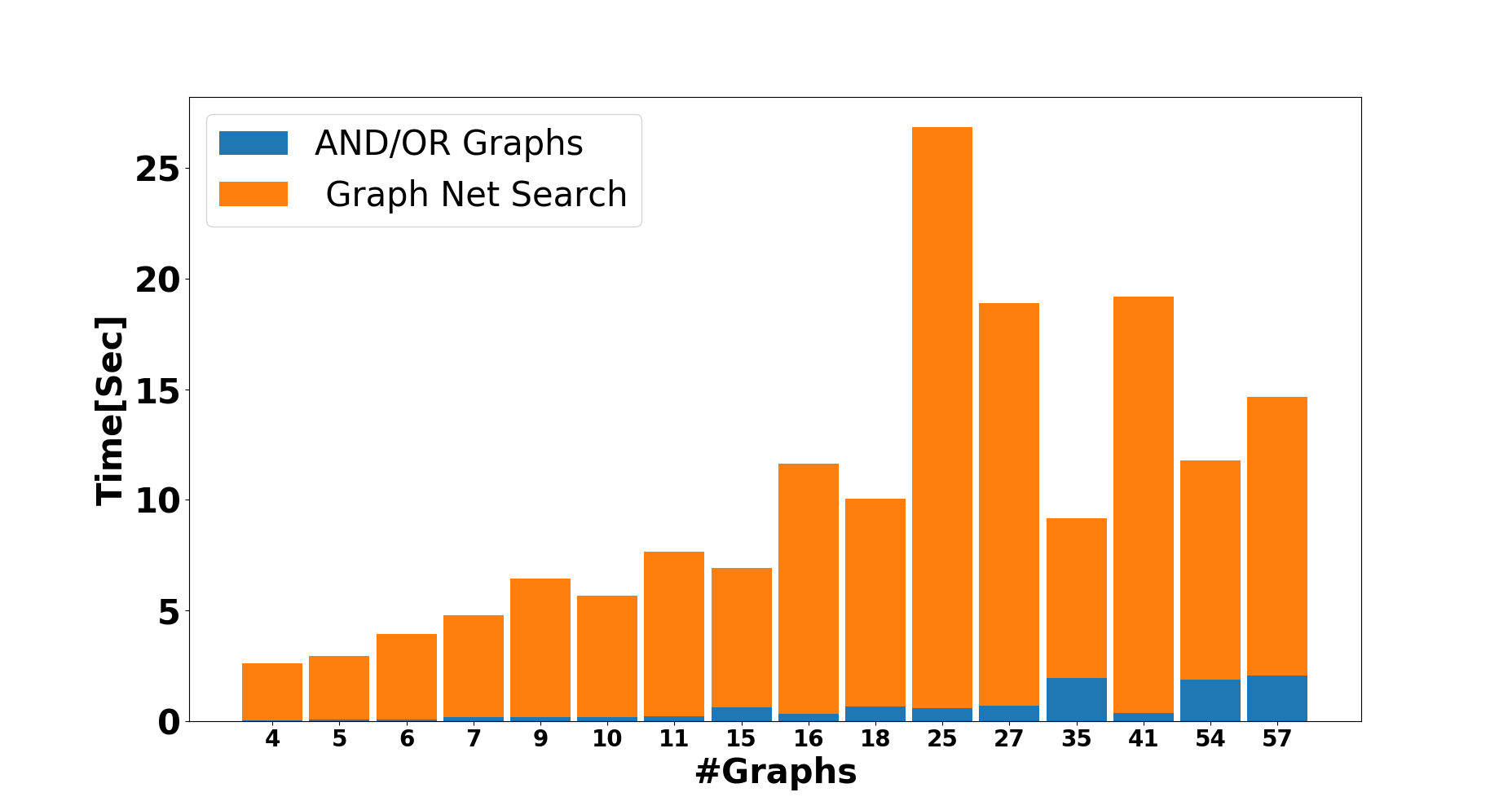}\label{fig:compa}}
\subfloat[]{\includegraphics[width=0.35\textwidth]{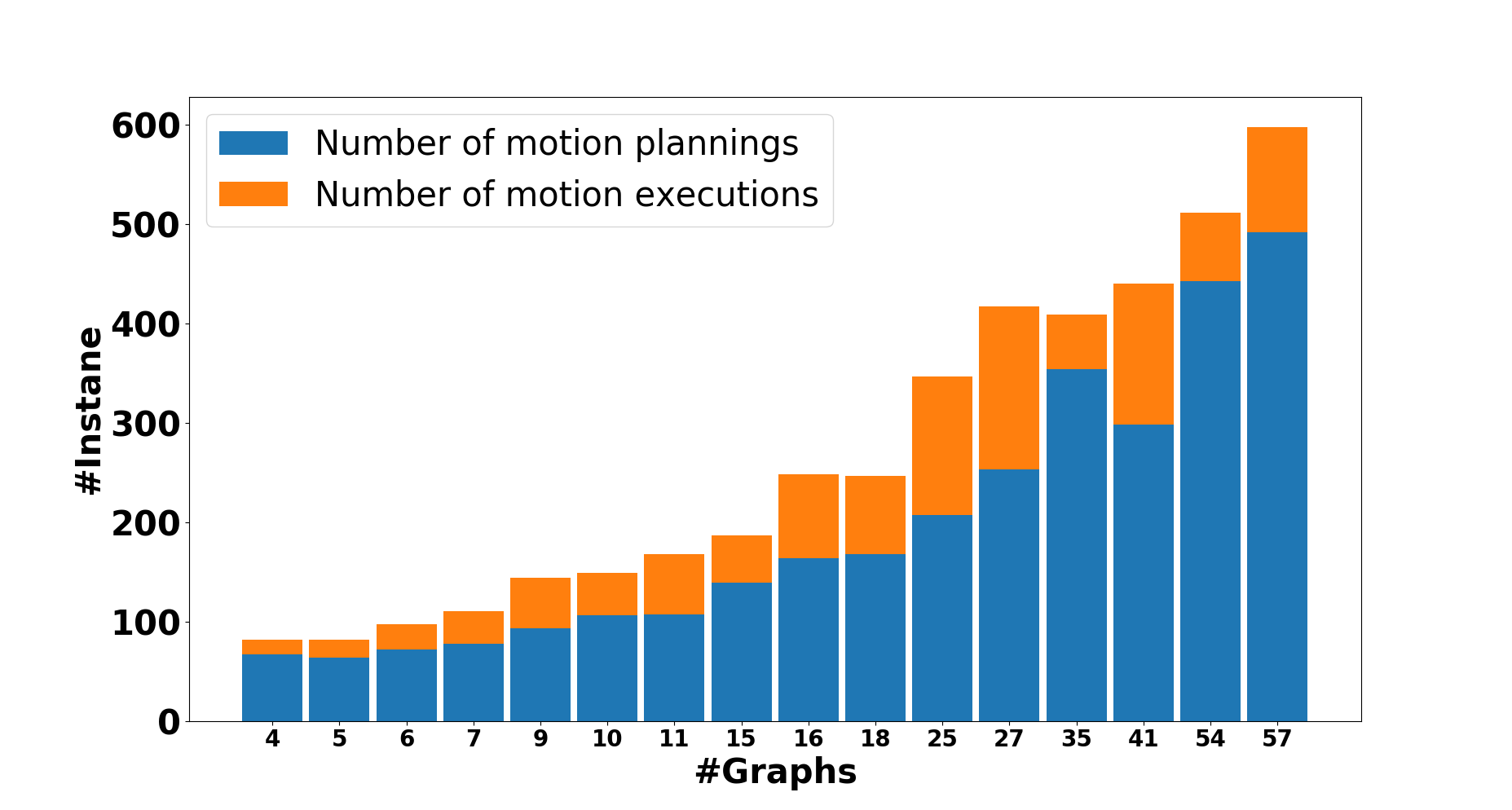}\label{fig:compb}}
\subfloat[]{\includegraphics[width=0.35\textwidth]{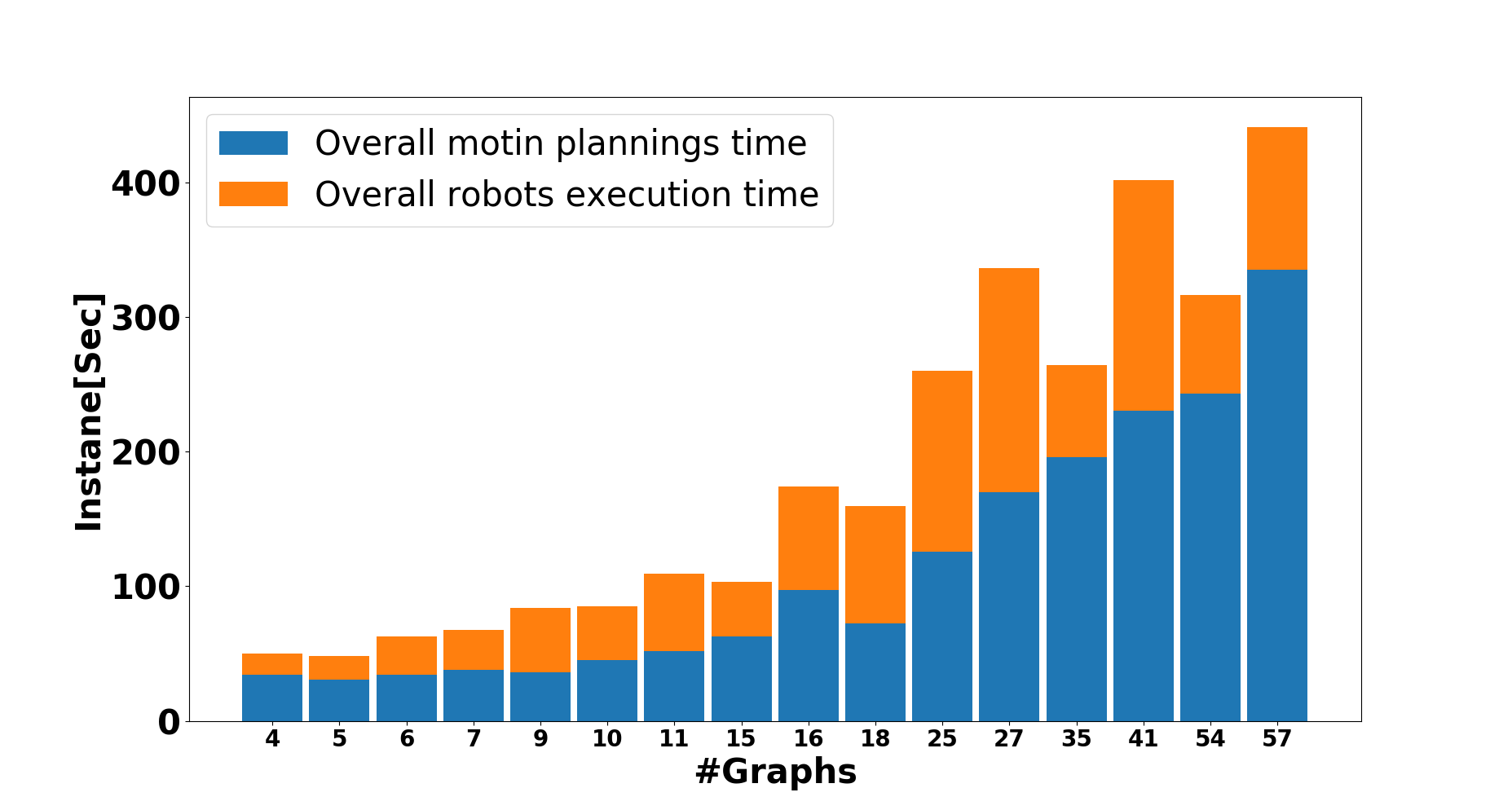}\label{fig:compc}}
\vspace{-0.1cm}
\caption{Various histograms with and increasing AND/OR graph network depth.}
\label{fig:comp1}
\end{figure}
We begin with 6 objects in the workspace with two of them being target objects. 
To test the scalability to an increasing number of objects, we perform experiments with up to 64 objects. 
For a given number of objects the experiment is conducted 3 times, and in each experiment the target objects are chosen randomly.
Table~\ref{tab:compana} reports the average combined planning and execution times for robots $r_1$ and $r_2$, focusing on the architecture's modules. 
AND/OR graph and graph network search times are per number of grown graphs. 
It must be noted that motion planning failures due to actuation errors or grasping failures lead to re-plans, and therefore to a larger number of motion planning attempts.

Table~\ref{tab:table1} report the average network depth $d$, the average total task planning time, the average total motion planning time, the average number of motion planning attempts, and the average number of objects to be re-arranged for robots $r_1$ and $r_2$, respectively. 
As the AND/OR graph network depth $d$ increases, task planning times are \textit{almost} linear with respect to $d$. 
This is so because for an AND/OR graph network with each graph consisting of $n$ nodes, the time complexity is only $O(nd)$~\cite{karami2021arxiv}. 
In contrast, PDDL-based planners are characterized by a search complexity of $O(n \log n)$, where $n \approx 2^{13}$ for the table-top scenario. 
\begin{table}[t!]
\centering
\scalebox{0.72}{
\begin{tabular}{|c|c|c|c|c|c|c|c|c|c|} 
 \cline{3-10}

\multicolumn{2}{c}{} & \multicolumn{4}{|c}{$r_1$} & \multicolumn{4}{|c|}{$r_2$}\\
 \hline
Objects & d     & TP [s]    & MP [s]    & MP attempts   & Objects re-arranged  & TP [s]    & MP [s]    & MP attempts   & Objects re-arranged \\
\hline
\hline
6       & 2.33  & 1.5505    & 16.339    & 36.66         & 1.66 & 1.1201    & 13.207    & 31.33         & 1.0 \\
\hline 
8       & 2.66  & 1.422     & 13.721    & 32.66         & 1.66& 1.1478    & 18.154    & 28.66         & 1.0 \\
\hline
9       & 2.33  & 1.3532    & 16.909    & 30.66         & 1.0 & 2.415     & 18.520    & 40.0          & 2.33\\
\hline
12      & 4     & 2.4025    & 20.3363   & 42.66         & 2.66 & 2.0532    & 17.825    & 37.66         & 2.33 \\
\hline 
16      & 4.66  & 2.755     & 20.1718   & 47.66         & 3.33 & 2.6326    & 22.830    & 45.66         & 3.33\\
\hline
20      & 3.66  & 2.125     & 17.618    & 42.33         & 2.66& 4.022     & 25.182    & 61.33         & 5.33  \\
\hline
30      & 5.75  & 3.464     & 25.854    & 56.75         & 4.25 & 4.748     & 45.226    & 83.75         & 5.75 \\
\hline
49      & 17    & 8.642     & 81.361    & 147.0         & 6.33 & 6.886     & 49.593    & 106.66        & 10.66\\
\hline
64      & 22.6  & 9.053     & 132.179   & 203.0         & 12.2 & 5.271     & 66.994    & 109.875       &7.0 \\
\hline
\end{tabular}}
\caption{Statistics for robots $r_1$ and $r_2$. Legenda: $d$ - average AND/OR graph network depth, TP - average total task planning time, MP - average total motion planning time.}
\label{tab:table1}
\end{table}
Fig.~\ref{fig:comp1} shows different histograms for an increasing network depth $d$. 
In particular, Fig.~\ref{fig:comp1}(a) shows the total task planning time with $d$. 
One can readily observes that the linearity in planning time is not strictly followed. 
For example, the time for $d=25$ is greater than the time for $d=41$. 
In many cases, due to motion planning failures a new graph is expanded before reaching the terminal node, which implies that more nodes are traversed for $d=25$ compared to $d=41$, therefore explaining the variations. 
Fig.~\ref{fig:comp1}(b) plots the number of motion planning attempts and the total executions with an increasing $d$. 
An increase in $d$ in most cases correspond to a higher degree of clutter. 
Therefore, as depth $d$ or the graphs increase the motion planning attempts increases as well. 
However, motion planning failures can also increase the depth $d$ since a new graph need to be expanded. 
This explains the slight deviation in the trend. 
Fig.~\ref{fig:comp1}(c) reports the total motion planning and execution times with an increasing $d$, and the plot readily follows from the discussions above. 

\vspace{-0.3cm}
\section{Conclusion}
\label{sec:conclusion}
\vspace{-0.2cm}
We present a task allocation approach for integrated multi-robot task and motion planning capable of handling tasks with an unknown number of sub-tasks. 
Our obstacles selection strategy combined with the introduced utility function allocates task to the available robots. 
Allocated tasks are then achieved by our TMP method, which comprises an AND/OR graph network based task planner that is robust to the issue of an unknown number of sub-tasks. 
We show that our method is capable of handling varying degree of clutter, i.e., the number of objects in the workspace. Currently, the objects hindering each tasks are removed from the workspace. 
 As a result, subsequent tasks need not consider theses objects while computing the utility. 
 There may be situations in which removing objects from the workspace will not be possible, and the planner will have to select free space for object placement. Future work will seek to develop a non-myopic technique to predict possible future trajectories based on the previous ones and the current environment configuration so as to enable the selection of smart object placements.

\bibliographystyle{splncs04}
\bibliography{References}

\begin{thebibliography}{10}
\providecommand{\url}[1]{\texttt{#1}}
\providecommand{\urlprefix}{URL }
\providecommand{\doi}[1]{https://doi.org/#1}

\bibitem{basile2012RCIM}
Basile, F., Caccavale, F., Chiacchio, P., Coppola, J., Curatella, C.:
  Task-oriented motion planning for multi-arm robotic systems. Robotics and
  Computer-Integrated Manufacturing  \textbf{28}(5),  569--582 (2012)

\bibitem{bertolucci2021TPLP}
Bertolucci, R., Capitanelli, A., Dodaro, C., Leone, N., Maratea, M.,
  Mastrogiovanni, F., Vallati, M.: Manipulation of articulated objects using
  dual-arm robots via answer set programming. Theory and Practice of Logic
  Programming  \textbf{21}(3),  372--401 (2021)

\bibitem{chang1971AI}
Chang, C.L., Slagle, J.R.: {An admissible and optimal algorithm for searching
  AND/OR graphs}. Artificial Intelligence  \textbf{2}(2),  117--128 (1971)

\bibitem{dantam2016RSS}
Dantam, N.T., Kingston, Z.K., Chaudhuri, S., Kavraki, L.E.: Incremental task
  and motion planning: A constraint-based approach. In: Robotics: Science and
  Systems (2016)

\bibitem{darvish2020hierarchical}
Darvish, K., Simetti, E., Mastrogiovanni, F., Casalino, G.: A hierarchical
  architecture for human--robot cooperation processes. IEEE Transactions on
  Robotics  \textbf{, to appear} (2020)

\bibitem{dornhege2009ICAPS}
Dornhege, C., Eyerich, P., Keller, T., Tr{\"u}g, S., Brenner, M., Nebel, B.:
  {Semantic Attachments for Domain-Independent Planning Systems}. In:
  International Conference on Automated Planning and Scheduling (ICAPS). pp.
  114--121. Thessaloniki, Greece (September 2009)

\bibitem{dornhege2009SSRR}
Dornhege, C., Gissler, M., Teschner, M., Nebel, B.: Integrating symbolic and
  geometric planning for mobile manipulation. In: Safety, Security \& Rescue
  Robotics (SSRR), IEEE International Workshop on. pp.~1--6. IEEE (2009)

\bibitem{erdem2011ICRA}
Erdem, E., Haspalamutgil, K., Palaz, C., Patoglu, V., Uras, T.: Combining
  high-level causal reasoning with low-level geometric reasoning and motion
  planning for robotic manipulation. In: 2011 IEEE International Conference on
  Robotics and Automation. pp. 4575--4581. IEEE (2011)

\bibitem{garrett2018IJRR}
Garrett, C.R., Lozano-Perez, T., Kaelbling, L.P.: {FFRob: Leveraging symbolic
  planning for efficient task and motion planning}. The International Journal
  of Robotics Research  \textbf{37}(1),  104--136 (2018)

\bibitem{garrett2020ICAPS}
Garrett, C.R., Lozano-P{\'e}rez, T., Kaelbling, L.P.: Pddlstream: Integrating
  symbolic planners and blackbox samplers via optimistic adaptive planning. In:
  Proceedings of the International Conference on Automated Planning and
  Scheduling. vol.~30, pp. 440--448 (2020)

\bibitem{gerkey2004IJRR}
Gerkey, B.P., Matari{\'c}, M.J.: A formal analysis and taxonomy of task
  allocation in multi-robot systems. The International Journal of Robotics
  Research  \textbf{23}(9),  939--954 (2004)

\bibitem{ghallab2016book}
Ghallab, M., Nau, D., Traverso, P.: Automated planning and acting. Cambridge
  University Press (2016)

\bibitem{henkel2019IROS}
Henkel, C., Abbenseth, J., Toussaintl, M.: An optimal algorithm to solve the
  combined task allocation and path finding problem. In: 2019 IEEE/RSJ
  International Conference on Intelligent Robots and Systems (IROS). pp.
  4140--4146. IEEE (2019)

\bibitem{jiang2019IROS}
Jiang, Y., Yang, F., Zhang, S., Stone, P.: {Task-Motion Planning with
  Reinforcement Learning for Adaptable Mobile Service Robots}. In: IROS. pp.
  7529--7534 (2019)

\bibitem{kaelbling2013IJRR}
Kaelbling, L.P., Lozano-P{\'e}rez, T.: Integrated task and motion planning in
  belief space. The International Journal of Robotics Research
  \textbf{32}(9-10),  1194--1227 (2013)

\bibitem{karami2020task}
Karami, H., Darvish, K., Mastrogiovanni, F.: A task allocation approach for
  human-robot collaboration in product defects inspection scenarios. In: 2020
  29th IEEE International Conference on Robot and Human Interactive
  Communication (RO-MAN). pp. 1127--1134. IEEE

\bibitem{karami2021arxiv}
Karami, H., Thomas, A., Mastrogiovanni, F.: {A Task-Motion Planning Framework
  Using Iteratively Deepened AND/OR Graph Networks}. arXiv preprint
  arXiv:2104.01549  (2021)

\bibitem{korsah2013IJRR}
Korsah, G.A., Stentz, A., Dias, M.B.: A comprehensive taxonomy for multi-robot
  task allocation. The International Journal of Robotics Research
  \textbf{32}(12),  1495--1512 (2013)

\bibitem{kuffner2000ICRA}
Kuffner, J.J., LaValle, S.M.: Rrt-connect: An efficient approach to
  single-query path planning. In: Robotics and Automation, 2000. Proceedings.
  ICRA'00. IEEE International Conference on. vol.~2, pp. 995--1001. IEEE (2000)

\bibitem{lagriffoul2018RAL}
Lagriffoul, F., Dantam, N.T., Garrett, C., Akbari, A., Srivastava, S., Kavraki,
  L.E.: Platform-independent benchmarks for task and motion planning. Robotics
  and Automation Letters  (2018)

\bibitem{latombe1991robot}
Latombe, J.C.: Robot Motion Planning. Kluwer Academic Publishers (1991)

\bibitem{lo2018AAMAS}
Lo, S.Y., Zhang, S., Stone, P.: Petlon: Planning efficiently for
  task-level-optimal navigation. In: Proceedings of the 17th International
  Conference on Autonomous Agents and MultiAgent Systems. pp. 220--228.
  International Foundation for Autonomous Agents and Multiagent Systems (2018)

\bibitem{mcdermott1998AIPS}
McDermott, D., Ghallab, M., Howe, A., Knoblock, C., Ram, A., Veloso, M., Weld,
  D., Wilkins, D.: {PDDL- The Planning Domain Definition Language}. In: AIPS-98
  Planning Competition Committee (1998)

\bibitem{mirrazavi2018IJRR}
Mirrazavi~Salehian, S.S., Figueroa, N., Billard, A.: A unified framework for
  coordinated multi-arm motion planning. The International Journal of Robotics
  Research  \textbf{37}(10),  1205--1232 (2018)

\bibitem{motes2019RAL}
Motes, J., Sandstr{\"o}m, R., Adams, W., Ogunyale, T., Thomas, S., Amato, N.M.:
  Interaction templates for multi-robot systems. IEEE Robotics and Automation
  Letters  \textbf{4}(3),  2926--2933 (2019)

\bibitem{motes2020RAL}
Motes, J., Sandstr{\"o}m, R., Lee, H., Thomas, S., Amato, N.M.: Multi-robot
  task and motion planning with subtask dependencies. IEEE Robotics and
  Automation Letters  \textbf{5}(2),  3338--3345 (2020)

\bibitem{munoz2016RAS}
Mu{\~n}oz, P., R-Moreno, M.D., Barrero, D.F.: Unified framework for
  path-planning and task-planning for autonomous robots. Robotics and
  Autonomous Systems  \textbf{82},  1--14 (2016)

\bibitem{preda2015IROS}
Preda, N., Manurung, A., Lambercy, O., Gassert, R., Bonf{\`e}, M.: Motion
  planning for a multi-arm surgical robot using both sampling-based algorithms
  and motion primitives. In: 2015 IEEE/RSJ International Conference on
  Intelligent Robots and Systems (IROS). pp. 1422--1427. IEEE (2015)

\bibitem{rodriguez2016IROS}
Rodr{\'\i}guez, C., Su{\'a}rez, R.: Combining motion planning and task
  assignment for a dual-arm system. In: 2016 IEEE/RSJ International Conference
  on Intelligent Robots and Systems (IROS). pp. 4238--4243. IEEE (2016)

\bibitem{coppeliaSim}
Rohmer, E., Singh, S.P.N., Freese, M.: {CoppeliaSim (formerly V-REP): a
  Versatile and Scalable Robot Simulation Framework}. In: Proc. of The
  International Conference on Intelligent Robots and Systems (IROS) (2013),
  www.coppeliarobotics.com

\bibitem{sharon2015AI}
Sharon, G., Stern, R., Felner, A., Sturtevant, N.R.: Conflict-based search for
  optimal multi-agent pathfinding. Artificial Intelligence  \textbf{219},
  40--66 (2015)

\bibitem{srivastava2014ICRA}
Srivastava, S., Fang, E., Riano, L., Chitnis, R., Russell, S., Abbeel, P.:
  Combined task and motion planning through an extensible planner-independent
  interface layer. In: Robotics and Automation (ICRA), IEEE International
  Conference on. pp. 639--646. IEEE (2014)

\bibitem{stilman2008IJRR}
Stilman, M., Kuffner, J.: Planning among movable obstacles with artificial
  constraints. The International Journal of Robotics Research
  \textbf{27}(11-12),  1295--1307 (2008)

\bibitem{sucan2013moveit}
Sucan, I.A., Chitta, S.: Moveit! (2013)

\bibitem{sucan2012RAM}
{\c{S}}ucan, I.A., Moll, M., Kavraki, L.E.: The {O}pen {M}otion {P}lanning
  {L}ibrary. {IEEE} Robotics \& Automation Magazine  \textbf{19}(4),  72--82
  (December 2012). \doi{10.1109/MRA.2012.2205651},
  \url{https://ompl.kavrakilab.org}

\bibitem{thomas2020IRIM}
Thomas, A., Karami, H., Mastrogiovanni, F.: {Iterative AND/OR Graphs for
  Task-Motion Planning}. In: Italian Conference on Robotics and Intelligent
  Machines (I-RIM) (2020)

\bibitem{thomas2019ISRR}
Thomas, A., Mastrogiovanni, F., Baglietto, M.: {Task-Motion Planning for
  Navigation in Belief Space}. In: The International Symposium on Robotics
  Research (2019)

\bibitem{thomas2020STAIRS}
Thomas, A., Mastrogiovanni, F., Baglietto, M.: {Towards Multi-Robot Task-Motion
  Planning for Navigation in Belief Space}. In: European Starting AI
  Researchers’ Symposium. CEUR (2020)

\bibitem{toussaint2015IJCAI}
Toussaint, M.: Logic-geometric programming: An optimization-based approach to
  combined task and motion planning. In: Twenty-Fourth International Joint
  Conference on Artificial Intelligence (2015)

\bibitem{umay2019ROSE}
Umay, I., Fidan, B., Melek, W.: An integrated task and motion planning
  technique for multi-robot-systems. In: 2019 IEEE International Symposium on
  Robotic and Sensors Environments (ROSE). pp.~1--7. IEEE (2019)

\bibitem{zlot2006phd}
Zlot, R.: {An auction-based approach to complex task allocation for multirobot
  teams, PhD thesis}. Robotics Institute, Carnegie Mellon University, 5000
  Forbes Ave.  (2006)

\end{thebibliography}


\end{document}